\documentclass[journal=ancham]{achemso}
\SectionNumbersOn 

\usepackage{chemformula} 
\usepackage[T1]{fontenc} 
\usepackage{pgf}
\usepackage{microtype}
\usepackage{graphicx}
\usepackage{subfigure}
\usepackage{multirow}
\usepackage{hyperref}
\usepackage{chngcntr}
\usepackage{booktabs, caption, makecell}
\usepackage{threeparttable}
\usepackage{algorithm}
\usepackage{algpseudocode}
\usepackage{amsmath,amssymb}
\usepackage{layouts}
\usepackage{setspace}

\author{Bo Li}
\email{blia@dtu.dk}

\author{Mikkel N. Schmidt}
\author{Tommy S. Alstrøm}
\affiliation[Compute]
{Department of Applied Mathematics and Computer Science, Technical University of Denmark, Kgs Lyngby, Denmark}

\title{Raman Spectrum Matching with Contrastive Representation Learning}

\abbreviations{RS, ML, DL, SERS}

\begin{document}

\begin{tocentry}
  \includegraphics[width=8cm]{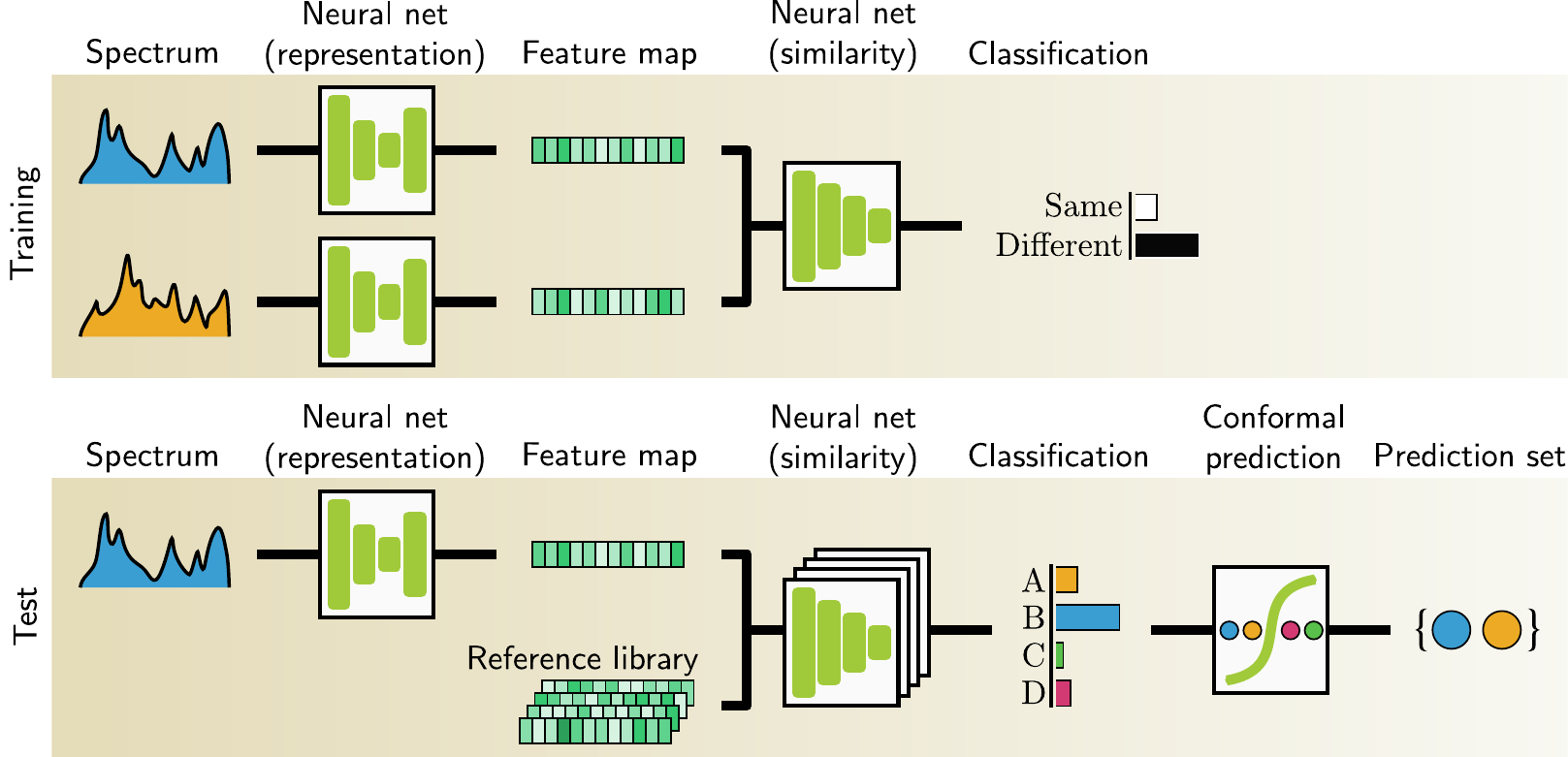}
    Raman spectrum matching with contrastive representation learning    




\end{tocentry}

\begin{abstract}
Raman spectroscopy is an effective, low-cost, non-intrusive technique often used for chemical identification.
Typical approaches are based on matching observations to a reference database, which requires careful preprocessing, or supervised machine learning, which requires a fairly large number of training observations from each class. 
We propose a new machine learning technique for Raman spectrum matching, based on contrastive representation learning, that requires no preprocessing and works with as little as a single reference spectrum from each class.
On three datasets we demonstrate that our approach significantly improves or is on par with the state of the art in prediction accuracy, and we show how to compute conformal prediction sets with specified frequentist coverage. 
Based on our findings, we believe contrastive representation learning is a promising alternative to existing methods for Raman spectrum matching.
\end{abstract}



\section{Introduction}
Identification of unknown substances is critical in many fields such as food product quality control~\cite{AMJAD2018124}, pharmaceutical drugs characterization~\cite{lussier2020a}, forensic investigation~\cite{huang2021a}, and bacteria detection~\cite{Ho2019RapidIO}. Raman spectroscopy is often used in these applications~\cite{lussier2020a, Ho2019RapidIO, huang2021a} as a low-cost, non-invasive technique. Raman spectroscopy provides a chemical fingerprint by using scattered light to probe the vibrational modes of a sample~\cite{lussier2020a}. When photons from a laser interact with the molecule, the scattered photons are shifted in frequency, and a spectrum is measured containing peaks representative of the chemical composition of the substance~\cite{LIU2019175}. 

Substance identification is then usually carried out by matching a measured spectrum to a reference library. This process can be automated by computing a similarity score between the acquired spectrum and each spectra in the library~\cite{Lafuente2016ThePO}. The computation of the similarity scores usually involves preprocessing the spectra in terms of baseline correction, smoothing, and normalization~\cite{C7AN01371J, doi:10.1021/ac5026368}. 
Matrix factorization methods such as multivariate curve resolution with non-negativity constraints (also known as non-negative matrix factorization) are often applied to extract features from the spectra~\cite{C4AY00571F, 6958925, doi:10.1021/ac901350a}. These features are then commonly used as input to machine learning classifiers such as support vector machines (SVM), random forest (RF)~\cite{doi:10.1021/acs.jpclett.9b02517}, K nearest neighbor (KNN)~\cite{lussier2020a} classifiers, or tree-based method~\cite{SEVETLIDIS2019121, doi:10.1021/acs.analchem.0c04576}. 

Appropriate preprocessing is essential for these methods to work well in practice, and their design requires extensive domain knowledge~\cite{C7AN01371J}. Handling the preprocessing and model training as separate steps may result in inadvertently removing useful spectral information~\cite{LIU2019175}. Additionally, these approaches may fail when dealing with complex spectra~\cite{huang2021a, C7AN01371J, FUKUHARA201911}. To alleviate this problem, we investigate how deep neural networks can be used to jointly learn a feature representation and a classifier by optimizing the entire processing end-to-end.

Deep neural networks have achieved superior performance across many fields~\cite{DBLP:conf/cvpr/Chollet17, DBLP:conf/cvpr/HeZRS16, DBLP:conf/nips/Lakshminarayanan17, lussier2020a}. In the context of spectral matching, neural networks are usually trained in a supervised manner, optimized directly to map an observed spectrum to a score for each reference substance in the training database~\cite{LIU2019175, C7AN01371J, Ho2019RapidIO, doi:10.1021/acs.analchem.1c00431}. Typically a convolutional neural network (CNN) is used since it can detect similarities in peaks and patterns across different wavenumbers~\cite{schmidt2019icassp}, leading to better sample efficiency. For example,~\citet{Ho2019RapidIO} identify bacteria with high accuracy on low signal-to-noise spectra using a ResNet~\cite{DBLP:conf/cvpr/HeZRS16}-based classifier. However, the performance of CNN-based supervised classifiers strongly depends on the amount of training data, and in many practical cases only a few spectra are available per substance~\cite{LIU2019175}. Additionally, if the reference database is modified, a supervised classifier needs to be retrained, which induces impractical computational costs~\cite{LIU2019175}. 

In use-cases such as detection of trace amounts in chemical threat monitoring, where responders need to decide on further actions, it is important that the classification uncertainty is appropriately characterized. When the output of a spectrum matching algorithm is given as a probability of detection, the score must be well-calibrated. It is well known that deep neural networks do not inherently output well-calibrated probabilities~\cite{Guo2017}. To ensure that the prediction scores are well-calibrated, further statistical methods need to be utilized. Reporting a set of candidate detections can instead be useful in cases where it is critical to discover the presence of a substance. In this way, the responders can screen for the existence of a target substance at a certain probability threshold.

In this paper, we present a method for precise, data-efficient, and uncertainty-calibrated classification of Raman spectra. Rather than training a supervised model to classify a spectrum among a set of classes, we train a neural network to predict if two spectra belong to the same class or not. This is done by processing two spectra by the same Siamese network, and computing a similarity score between their output feature representations. After training the network, a new spectrum can be classified by computing its representation and similarity score with reference spectra. This approach makes it easy to understand how the model works, as it supplies both a class label and the library spectra used in the match. A further benefit, not explored in this work, is that the reference database can be modified without retraining the model. Finally, we propose a conformal~\cite{angelopoulos2021uncertainty, DBLP:conf/nips/RomanoSC20} procedure to compute a prediction set that is guaranteed to include the true class with a high user-specified probability. This is demonstrated on a Surface Enhanced Raman Spectroscopy (SERS) dataset, where a large number of spectra are available per class.

Our proposed system includes several technical novelties, including 
\begin{itemize}
\item an optimized neural network architecture for neural feature representation learning using depth-wise separable convolutions,
\item a novel procedure for data augmentation that improves learning with even a single spectrum from each class,
\item a tailored conformal prediction set procedure with a frequentist guarantee of including the correct class prediction.
\end{itemize}

We experimentally validate our proposed approach on three publicly available datasets, and demonstrate that we match or outperform the existing state-of-the-art. Our model does not rely on data preprocessing, and in fact performs significantly better on raw data. Finally, we demonstrate that our conformal prediction sets in practice include the correct class label on held-out test data with approximately the prescribed frequency and can be tuned to provide near-perfect classification sets of moderate size.
\section{Methods and materials}
\label{sec:methodology}

\subsection{Sample description}
We demonstrate our system on three publicly available datasets (two Raman and one SERS) which we refer to as \emph{Mineral}, \emph{Organic}, and \emph{Bacteria}, summarized below and in Table~\ref{tab:dataset_statistics}.

\begin{figure}[tbp]
    \centering
    \input{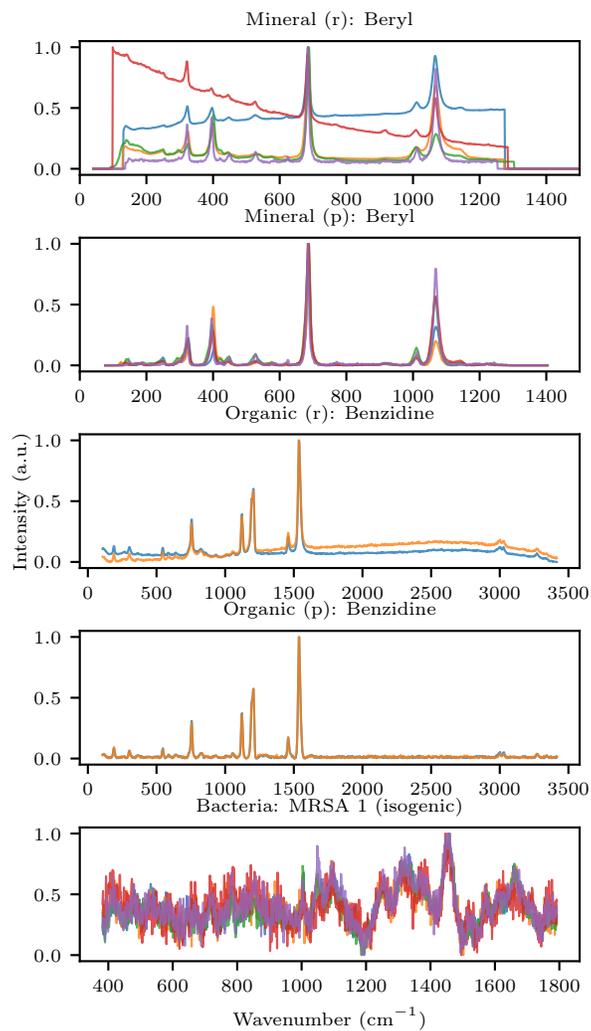}
    \caption{Example spectra from each dataset. The signal-to-noise ratio in the Bacteria dataset is substantially lower compared to the other datasets.}
    \label{fig:example_spectra}
\end{figure}

\begin{description}
\item[Mineral:] Minerals from the RRUFF~\cite{Lafuente2016ThePO} database. The spectra from each mineral are recorded under different measuring environments and with different wavelengths, and we use both the raw (r) preprocessed (p) spectra. 

\item[Organic:] Organic compounds collected and measured with several different excitation sources as described by \citet{organic_dataset}. We use both the raw (r) and preprocessed (p) spectra.

\item[Bacteria:] Bacterial Raman spectra described by \citet{Ho2019RapidIO}. This dataset consists of: reference dataset, reference-finetune dataset, and test dataset. The test dataset is measured following a similar experimental setup as reference-finetune dataset.

\begin{table*}[tbp]
\caption{Number of classes and spectra per dataset.}
\label{tab:dataset_statistics}
\begin{tabular}{lccccc}
\toprule
& Mineral (r) & Mineral (p) & Organic (r/p) & Bacteria \\
\midrule
Number of classes       & 487      & 1\,310           & 72             & 30       \\
Total number of spectra       & 1\,700     & 4\,864            & 216          & 66\,000   \\
Number of spectra per class & 2--40  & 2--40         & 3            & 2\,200    \\ 
\bottomrule
\end{tabular}
\end{table*}
\end{description}
Fig.~\ref{fig:example_spectra} shows a few example spectra from each dataset. Compared to other datasets, the weak Raman signal from bacterial cells result in low signal-to-noise (SNR) spectra, which poses challenges for achieving high identification accuracies. For additional details, see S.1 in the supplement.

\subsection{Model architecture}
We propose a contrastive representation learning method based on a Siamese neural network~\cite{Koch2015SiameseNN}. In this approach, the multiclass classification problem is re-cast as binary-classification by organizing the spectra into pairs and predicting whether the paired spectra are from the same or different classes, referred to as a positive or negative pair. This classification is achieved by first mapping the two spectra into learned feature representations using an identical representation network for each spectrum. Next, a second neural network is used to compute a similarity measure in the feature domain, resulting in a similarity score.

\subsubsection{Representation network}
The goal of the representation network is to extract spectral features that are useful for contrasting molecular structures. This includes capturing important characteristics such as locations and shapes of spectral peaks. Typical convolutional neural networks (CNNs), which extract features by repeated blocks of convolutions (multi-channel feature extraction) and pooling (dimensionality reduction), are not optimally suited because the successive pooling operations risk loosing the exact peak location information. Without pooling, however, the number of parameters in the neural network will increase drastically, and while this makes the model more expressive it also demands more training data.

As a compromise between expressivity and data efficiency, we propose to incorporate Xception~\cite{DBLP:conf/cvpr/Chollet17, DBLP:conf/aaai/SzegedyIVA17, DBLP:conf/cvpr/SzegedyLJSRAEVR15}, also known as depthwise seperable convolutions. This allows the neural net learn richer and more representative features while keeping the number of parameters low. Xception assumes that the the cross-channel and cross-wavenumber correlations are decoupled and thus can be mapped separately: The convolutional operation is factorized into a point-wise convolution along the feature channel dimension followed by a channel-wise convolution along the wavenumber dimension.

\begin{figure*}[tbp]
    \centering
    \includegraphics[width=.8\textwidth]{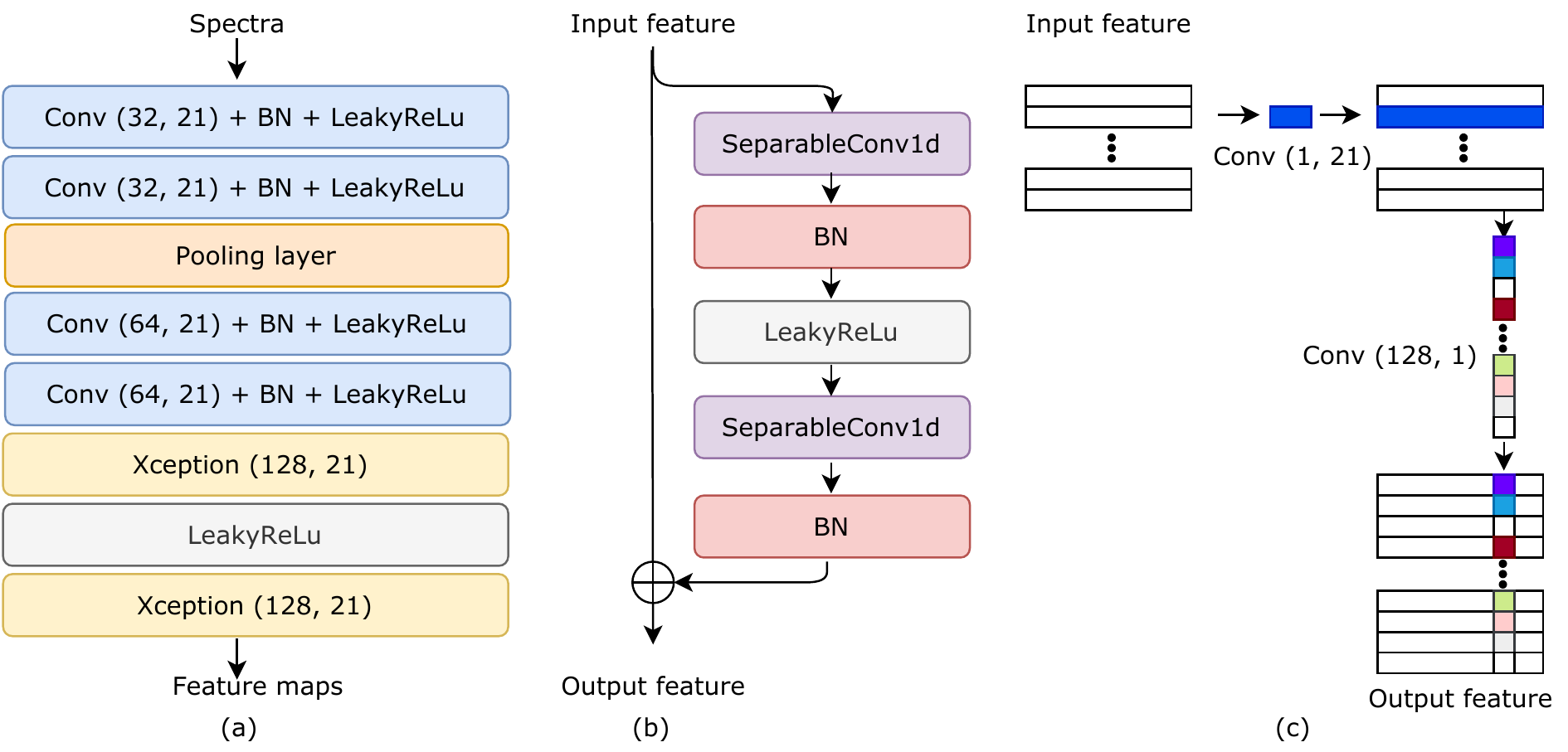}
    \caption{Model architecture. a) The full model architecture. b) An Xception block. c) A SeparableConv1d layer. The numbers in the bracket corresponds to the number of output channels and kernel size. We use the same architecture for all the datasets.}
    \label{fig:sec_2_model_architecture}
\end{figure*}

The architecture of our model can be seen in Fig.~\ref{fig:sec_2_model_architecture}.
Since depthwise separable convolutions are most effective in high dimensions~\cite{DBLP:conf/cvpr/Chollet17, DBLP:conf/aaai/SzegedyIVA17}, we first use two conventional convolutional blocks to transform the one-dimensional input into a larger number of channels. Each convolutional block contains two times of one-dimensional convolutional layers, batch normalization layers, and LeakyReLu layers. We use only a single pooling layer along the wavenumber dimension to void losing accurate peak location information. We then stack another two \textit{Xception} blocks on top as shown in Fig.~\ref{fig:sec_2_model_architecture}. We use residual connections~\cite{DBLP:conf/cvpr/HeZRS16} in the \textit{Xception} block since they have been shown to be essential in helping with convergence in terms of both speed and accuracy~\cite{DBLP:conf/cvpr/HeZRS16, DBLP:conf/aaai/SzegedyIVA17, DBLP:conf/cvpr/Chollet17}. Each block ends with a dropout layer to alleviate the risk of over-fitting the model. The produced feature maps from the last \textit{Xception} block are used as the final representation.

\subsubsection{Similarity network}
\begin{figure*}
\centering
\resizebox{1.0\textwidth}{!}
{\begin{minipage}{\textwidth}
\includegraphics[width = 1.0\textwidth]{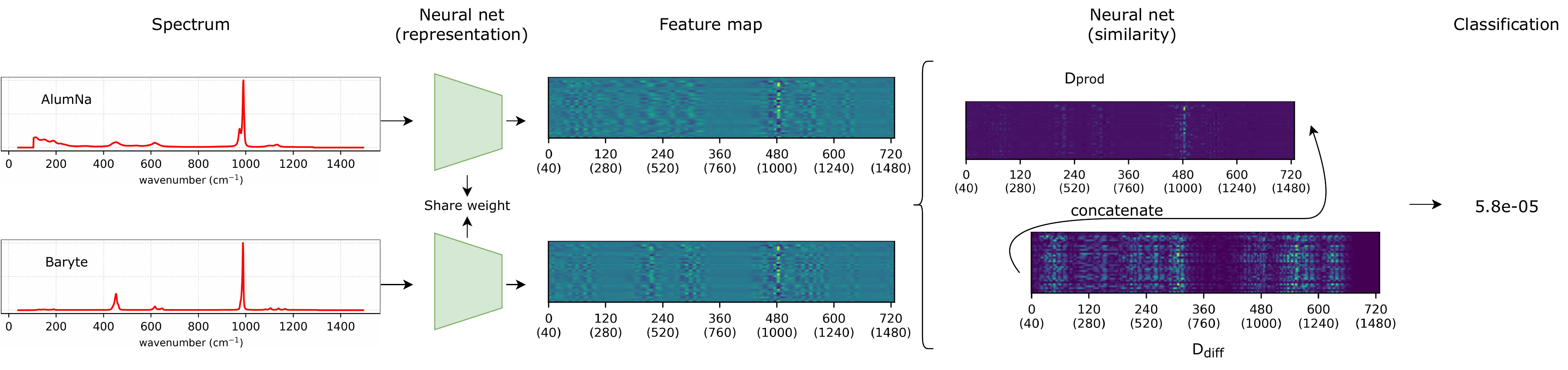}
\end{minipage}}
\caption{The proposed spectra matching approach. During training, we first extract representative features from the input spectra using a deep neural network. We then quantify the similarity by calculating the distance between the extracted feature maps using both element-wise multiplication ($D_{\text{prod.}}$) and absolute element-wise difference ($D_{\text{diff.}}$). The values in the bracket in the x-axis correspond to the wavenumber index in the input spectrum. These two distance maps are then concatenated along the wavenumber dimension and used to produce a single similarity score $p\in(0,1)$. After training, all spectra are allocated to the reference library, and at inference time, we calculate the similarity score between the unknown spectrum and each reference match by the similarity score.}
\label{fig:main_architecture}
\end{figure*}

To quantify the similarity of two spectra, we first calculate the element-wise product and absolute difference between their feature maps $F_1$ and $F_2$,
\begin{subequations}
    \begin{equation}
    D_{\text{prod.}}(c, w) = F_1(c, w) \cdot F_2(c, w),
    \label{eq:dot_product}
    \end{equation}
    \begin{equation}
    D_{\text{diff.}}(c, w) = |F_1(c, w) - F_2(c, w)|,
    \label{eq:l1_norm}
    \end{equation}
\end{subequations}

where $c$ and $w$ are the channel and wavenumber feature index. This results in two distance maps $D_\text{prod.}$ and $D_\text{diff.}$. Akin to dot product/cosine similarity, $D_\text{prod.}$ will be large when two spectra have strong features in common at the same wavenumber, whereas $D_\text{diff.}$ will be small where the two spectra have similar features irrespective of their magnitude. We concatenate the two distance maps along the wavenumber dimension and pass the resulting matrix to a 1D convolutional layer (kernel size 1, stride 1) and a fully connected layer followed by a sigmoid activation to yield the final similarity score $p\in (0,1)$. In Fig.~\ref{fig:main_architecture} we illustrate our framework using a pair of example spectra as input.

\subsubsection{Similarity score matching}
At inference time, we first calculate the similarity score between a test spectrum and each of the spectra in the reference library. We then group these similarity scores based on the reference label and take the maximum similarity score within each group. For the RRUFF and Organic dataset, we simply classify the test spectrum to the class that has the highest similarity score. For the Bacteria dataset, since the signal-to-noise ratio is relatively low and some bacteria could be quite similiar, we perform a top-$M$ majority voting strategy. This $M$ is experimentally optimized using the validation dataset.

\subsection{Model training}
Given a paired set of input spectra and the corresponding labels, we train the model using a cross-entropy objective with an L$_2$ regularization term
\begin{equation}
    L = -\frac{1}{N}\sum_{i=1}^{N} y_i \log p_i + (1 - y_i) \log (1 - p_i) + \lambda ||\theta||^2,
\end{equation}
where $N$ is the number of pairs per batch, $y_i$ is the label for the pair $i$ ($y_i=1$ for a positive pair and $y_i=0$ for a negative pair), and $p_i$ is the similarity score for pair $i$. The L$_2$ regularization term is included to alleviate over-fitting, with $\lambda=10^{-3}$ chosen experimentally. We train the model with Adam Optimizer with the initial learning of $5\cdot 10^{-5}$, which is decayed following cosine annealing schedule~\cite{DBLP:conf/iclr/GotmareKXS19}. During training, the dropout rate in each dropout layer is set to be 0.5~\cite{C7AN01371J}, and we deactivate dropout layers during testing. Once the training is done, we allocate all the learned representations from the training data to the reference library. 

\subsubsection{Data augmentation}
\begin{figure}[tbp]
\centering
\input{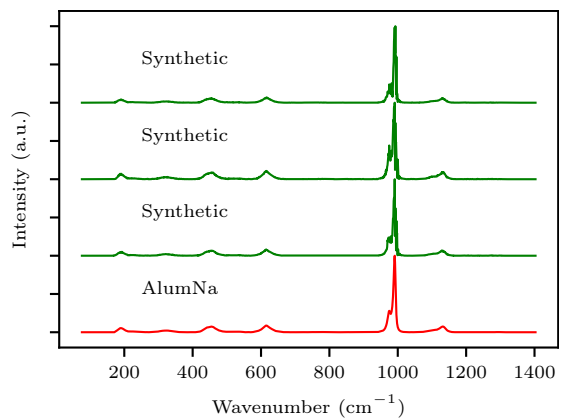}
\caption{An example of the original and synthetic spectra. We add more noise on the locations which are closer to peaks. The simulated spectra are similar to the measurement} 
\label{fig:augmentation_example}
\end{figure}

Since the size of the dataset strongly influences the performance of deep neural networks~\cite{C7AN01371J}, it is common practice to augment the data, for example by adding slightly modified copies of existing data. We propose to perform different augmentation techniques depending on the number of available spectra per class:
\begin{description}
    \item[>100:] If a class has more than 100 spectra, we do not perform any augmentation as sampling more spectra tends to increase the within-class variation even further.
    \item [2--100:] If a class has between 2 and 100 spectra, we add linear combinations between pairs of spectra with randomly simulated coefficients to generate new spectra following~\citet{C7AN01371J}.
    \item[1:] If a class has only one spectrum, linear combinations cannot be used. Instead, we augment the spectrum by adding noise simulated to mimic realistic spectral variation with highest variance near spectral peaks. We use Gaussian noise with amplitude proportional to the standard deviation $\sigma_w$ over a length $K$ sliding window of the first difference $d_w = s_w - s_{w - 1}$ of the intensities, 
    \begin{align}
        \mu_w &= \frac{1}{K}\sum_{k=0}^{K-1}d_{w+k}, &
        \sigma_w^2 &= \frac{1}{K}\sum_{k=0}^{K-1}(d_{w+k}-\mu_w)^2, &
        \hat{s}_w &\sim \mathcal{N}(s_w, \kappa\cdot\sigma_w^2),
    \end{align}
    where $s_w$ and $\hat s_w$ are the original and augmented spectra, and $w$ is the wavenumber index. The first difference is used to indicate the existence of peaks, and the standard deviation $\sigma_{w}$ can highlight the size and the visibilities of the peaks. The parameters $K=10$ and $\kappa=5$ were chosen experimentally to match typical spectral variation. An example of an original and three augmented spectra is shown in Fig~\ref{fig:augmentation_example}. We also experimented with more simple augmentation techniques which performed less well, see S.2 in the supplement. 
\end{description}

\subsubsection{Balancing positive and negative pairs}
One of the common issues associated with training a Siamese network is how to sample batches of positive and negative pairs such that the model efficiently learns both to distinguish dissimilar and match similar pairs. This is especially important in the application of spectral matching where there are typically hundreds to thousands of different classes, but the number of spectra per class can be very different (e.g., the majority of the minerals only have one spectrum).

If we have $N$ classes and each class has $M$ spectra on average, then the ratio between the number of possible positive and negative pairs is approximately $\frac{M-1}{M(N-1)}\ll 1$. As the number of negative pairs is far greater than the number of positive pairs, sampling all pairs uniformly may limit the model's capacity, so we use a fixed ratio $\beta$ of positive and negative pairs per batch sampled randomly at each iteration. This ratio is fixed for all the iterations per experiment and is chosen experimentally. We let $\beta=1$ for the Mineral dataset and $\beta=0.5$ for the Organic and Bacteria dataset (see S.3 in the supplement for details).

\subsubsection{Ensembling}
Rather than using a single model, we use an ensemble of models both to improve the predictive performance and to quantify the predictive uncertainty~\cite{DBLP:conf/mcs/Dietterich00, DBLP:conf/nips/Lakshminarayanan17,DBLP:conf/mlcw/CandelaRSBS05}. Following~\citet{DBLP:journals/corr/LeePCCB15}, we train these models with the same training data but with different random initialization. We then use the average predictions from the ensemble to make decisions for an unknown spectrum at test time. 

\subsection{Cross-validation}
We use a nested cross-validation that divides the data into three sets: a training set used to fit the neural network models, a validation set to optimize model hyperparameters, and a test set to give an unbiased evaluation of the performance.

\subsubsection{Test set}
For the Mineral and Organic datasets, because the number of spectra per class is limited, we randomly select one spectrum per class to form the test set and use the remaining spectra as the training/validation set. We perform this randomized leave-one-out four times to generate different data splits to demonstrate the robustness of the model. As for the Bacteria dataset, we simply follow~\citep{Ho2019RapidIO} and use the existing test dataset to demonstrate the performance. 

\subsubsection{Validation set}
The remaining data, not used for the test set, is used to form the training and validation sets. For the Mineral and Organic dataset, since many classes have only a few observed spectra, it is not data-efficient to hold out observations for hyperparameter optimization. Therefore, instead of selecting a small set from the training dataset to formulate the validation pairs, we propose to simulate spectra by adding noise onto the measured spectra and randomly shifting them w.r.t. wavenumber. For the positive pairs, we shift the spectra either up or down with a very small wavenumber, $3\  \text{cm}^{-1}$, as this is a common augmentation technique to simulate more spectra~\cite{C7AN01371J, LIU2019175}. For the negative pairs, we shift the spectrum either up or down with a larger random wavenumber, following a normal distribution with mean $150\ \text{cm}^{-1}$ and standard deviation $50\ \text{cm}^{-1}$. Since the scale of the shifting is relatively large, the peaks in the original and simulated spectra are at different locations and can thus serve as negative pairs. We simulate a new set of shifting parameters at each validation step to increase the diversity of the validation pairs. For the Bacteria dataset which is relatively large, we simply select 20 spectra (10 from the reference dataset and 10 from the reference-finetune dataset) from each class to formulate the validation positive and negative pairs. 

\subsection{Conformal prediction}

Understanding the relationship between matching accuracy and uncertainty level helps us know when the model is likely to make mistakes; however, in practice we need to make decisions based on the model output, and simply knowing when the model is likely to make mistakes is often not enough. Using the method of conformal prediction~\cite{Vovk20051, DBLP:journals/jmlr/ShaferV08, DBLP:conf/ecml/PapadopoulosPVG02, DBLP:conf/ictai/PapadopoulosVG07}, instead of outputting a single prediction, we can produce prediction set of varying size, that provably cover the true class with a high user-specified probability (i.e., 99\%).

Let $X_n$ denote a data point and $Y_n$ the corresponding true label. Conformal prediction~\cite{angelopoulos2021uncertainty, DBLP:conf/nips/RomanoSC20} aims to construct a predictive set $\mathcal{C}_{n, \alpha}(X_n)$ such that:
\begin{equation}
    \mathrm{P}(Y_{n} \subseteq \mathcal{C}_{n, \alpha}(X_{n})) \geq 1 - \alpha,
\end{equation}
where $\alpha\in (0, 1)$ specifies the desired coverage level. To find the predictive set $\mathcal{C}$, the procedure uses a threshold $\tau$ on the similarity scores to determine which class predictions should be included. This threshold is estimated using independently, identically distributed hold-out samples~\cite{angelopoulos2021uncertainty}, and in our setting we use the validation data. Since for the Mineral and Organic experiments the validation data is not independent (it was created by data augmentation) we limit our analysis to the Bacteria dataset. 

To find the threshold $\tau$ given the desired coverage level $\alpha$, we first compute the maximum similarity score for the correct class on the validation data. We then select $\tau$ as $\alpha_{\text{th}}$ quantile, such that the $1-\alpha$ proportion of the validation data have a similarity score greater than $\tau$. From the assumption that the validation and test data are identically distributed, we then have that confidence sets computed using $\tau$ to threshold the similarity score will have the desired coverage.

To prevent producing an empty predictive set in the cases where none of the predictions have probabilities that are higher than the threshold $\tau$, we always include the top-1 prediction in $\mathcal{C}$. To quantitatively evaluate the performance of conformal prediction, we calculate the empirical coverage, which measures the percentage of having the true class in the predictive set $\mathcal{C}$.

\section{Results and discussion}
We first quantitatively and qualitatively evaluate the predictive performance of the proposed method on all the datasets. Next, we demonstrate the relation between the predicted similarity score and matching accuracy. Finally, we quantitatively show the benefit of producing adaptive prediction sets using conformal prediction on the Bacteria dataset.

\subsection{Spectrum identification accuracy}
We compute the global average matching accuracy as the proportion of test spectra that are correctly identified, according to the similarity scores averaged across the ensemble. We compare our method with one-nearest-neighbors (1NN) baseline classifier and the existing state-of-the-art deep learning-based approaches. For the 1NN baseline we include Euclidean distance, Manhattan distance, and cosine similarity, as these similarity measures are widely used in commercial software for spectrum matching~\cite{LIU2019175}. Table~\ref{tab:overall_performance} displays our results together with a 95\% confidence interval across the five data splits, computed as $\pm 1.96\tfrac{\sigma_a}{\sqrt{4}}$ where $\sigma_a$ is the standard deviation of the accuracy.

In all cases our approach performs better or on par with the result in previous work. It is worth noting that the nearest neighbor (NN) relies on appropriate preprocessing steps~\cite{LIU2019175, organic_dataset} as they achieve much better performance on the preprocessed datasets. Contrary to that, our method performs best on the raw datasets which indicates that information useful for discriminating between classes may lost in the preprocessing. We observed that 1NN-cosine performs similar to our approach on the Organic (p) dataset. One of the reasons is that some classes (12 out of 72) in the preprocessed dataset contains virtually identical spectra, which could explain why the nearest neighbor and our reference matching approach strongly outperforms the existing work that uses supervised classification. We also achieved a much higher accuracy compared to the reported transferred learning accuracy that is obtained by using another bigger organic dataset as the base learner~\cite{organic_dataset}. 

\begin{table*}[tbp]
\caption{Performance together with a 95\% confidence interval over four different data splits on all the datasets (\%). The number of model parameters are given in parentheses. Our model significantly outperform the existing approaches on most data sets in terms of matching accuracy while using a much smaller model.}
\smallskip
\label{tab:overall_performance}
\begin{threeparttable}[t]
\centering
\begin{tabular}{@{}llllll@{}}
\toprule
 & Mineral (r) & Mineral (p) & Organic (r) & Organic (p) & Bacteria \\ \midrule
1NN Euclidean  & 34.0$\pm$0.9 & 78.5$\pm$0.5 & 72.9$\pm$3.5 & \textbf{97.7}$\pm$1.7 & 35.0$\pm$0.3 \\ \noalign{\smallskip}
\phantom{1NN} Manhattan  & 27.3$\pm$0.6& 80.0$\pm$0.6 & 62.3$\pm$2.8 & \textbf{94.9}$\pm$2.4 & 34.6$\pm$0.5 \\ \noalign{\smallskip}
\phantom{1NN} Cosine & 38.8$\pm$0.9 & 81.1$\pm$0.5 & 83.6$\pm$2.8 & \textbf{98.4}$\pm$1.3 & 36.3$\pm$0.7 \\ \noalign{\smallskip}
\midrule
Existing work & \begin{tabular}[c]{@{}c@{}}93.3$\pm$0.7\cite{C7AN01371J} \\(25.8M)\end{tabular} & \begin{tabular}[c]{@{}c@{}}88.4$\pm$0.5\cite{C7AN01371J} \\(0.19M)\end{tabular}  & - & \begin{tabular}[c]{@{}c@{}}92.6\cite{organic_dataset} \tnote{*} \\(27.5M)  \end{tabular} & \begin{tabular}[c]{@{}c@{}}\textbf{82.2}$\pm$0.3\cite{Ho2019RapidIO} \\(1.34M) \end{tabular} \\ \noalign{\smallskip}\midrule
Ours (0.23M) & \textbf{94.6}$\pm$0.3 & \textbf{91.7}$\pm$0.4  & \textbf{99.0}$\pm$0.6  & \textbf{97.5}$\pm$0.7 & \textbf{82.6}$\pm$0.5 \\ \noalign{\smallskip}
\bottomrule
\end{tabular}
\begin{tablenotes}\footnotesize
\item[*] Without applying transfer learning,~\citeauthor{organic_dataset} achieved 92.6 top-1 matching accuracy. After applying transfer learning, they achieved 96.4\% top-1 matching accuracy.
\end{tablenotes}
\end{threeparttable}
\end{table*}

\subsection{Qualitative evaluation}
\begin{figure*}[ht!]
    \centering
    \input{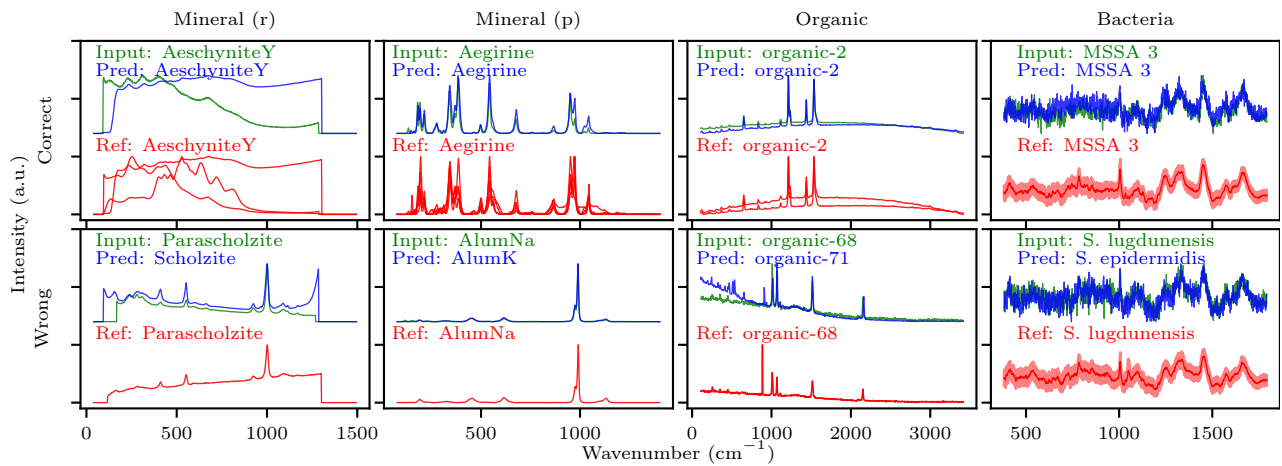}
    \caption{Examples of correctly and wrongly matched spectra for each dataset. \textit{Input} denotes the test spectrum, \textit{Pred} is the match retrieved from the reference library, and \textit{Ref} are all reference spectra from the correct class. For the Bacteria dataset, due to the large number of spectra per class \textit{Ref} shows the mean and standard deviation. In the failure cases, most of the input and retrieved spectra have similar chemical structure and thus their spectra also look similar.}
    \label{fig:qualitative_analysis}
\end{figure*}
One of the benefits of working with the Siamese network is that we can understand how the decision is made by looking at the matching regions/areas between the target and retrieved spectra as exemplified in Fig.~\ref{fig:qualitative_analysis}.

We notice that there are two scenarios where the model is most likely to match a spectrum wrong. First, it is difficult for the model to identify a spectrum correctly when it is very different from the ones in the reference library. For example, the spectra for mineral \textit{Parascholzite} in the reference library and test set look different due to different measuring wavelengths (785nm and 532nm). In addition, for bacteria \textit{S.lugdunensis}, the peak at around 1\,100 is not present in the input spectrum (see also S.5 in the supplementaty). These dissimilarities between the spectra from the same class are a limiting factor for the model. Secondly, if two classes have similar chemical compositions, it can also be hard for the model to distinguish between their spectra. For example, the spectra for the minerals \textit{AlumNa} and \textit{AlumK} are almost identical. However, most of the time, these classes tend to be matched wrong with the first candidate but correct with the second or third most probable candidates. Therefore, we next investigate how well we can construct predictive sets that are small for easy-to-recognize spectra but larger for hard-to-recognize spectra. 

\subsection{Uncertainty estimation}
\begin{figure}[ht!]
    \centering
    \input{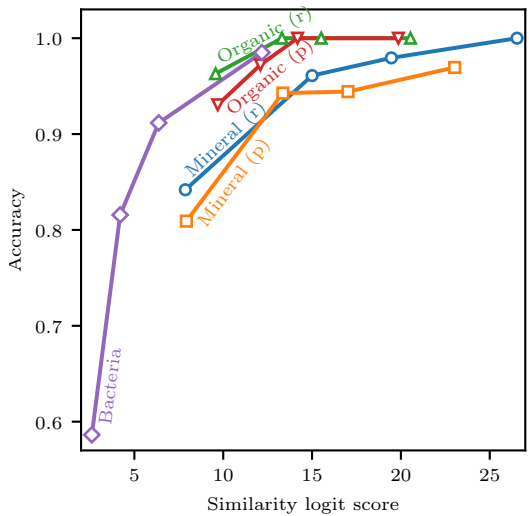}
    \caption{Percentage of the correct matching at different confidence levels on the test dataset (average over five different data splits). The higher the similarity logit score is, the more confident the model is. As the confidence increases, the percentage of correct matching also increases}
    \label{fig:uncertainty_distribution_on_test_dataset}
\end{figure}

To understand how well the predicted similarity score reflects the empirical error when evaluated on unseen dataset, we examine the relationship between the model predictions and matching accuracy. We group the predictions in intervals based on the quartiles of the predicted similarity logit score (network output before the sigmoid function) and calculate the average matching accuracy within each interval. The higher the logit score is, the more similar the model believes the input pair of spectra is. Fig.~\ref{fig:uncertainty_distribution_on_test_dataset} shows the result. We observe that as the similarity score increases, the matching accuracy also increases, which indicates that the similarity score is useful for quantifying the prediction uncertainty. 

\subsection{Variable-size prediction sets}
\begin{figure*}[ht!]
\centering
   \input{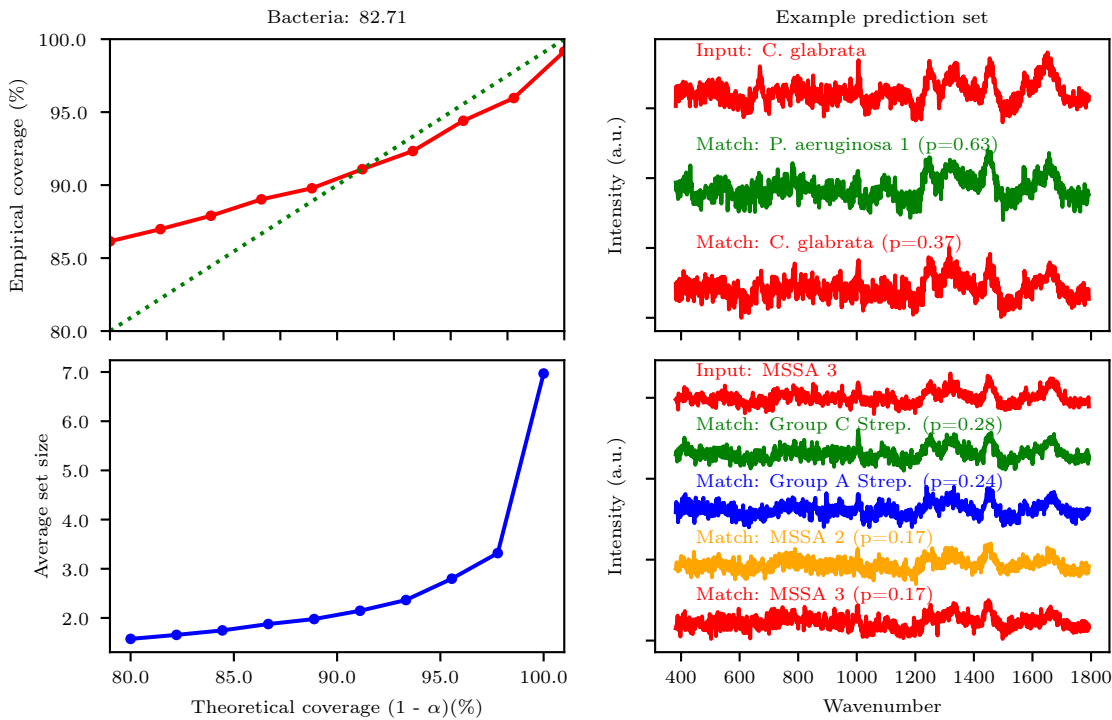}
   \caption[]{Left: empirical coverage, average set size and the theoretical coverage for Bacteria dataset. As the theoretical coverage increases, the empirical coverage also increase without drastically increasing the set size. Right: conformal prediction set examples with predicted matching probability (\textit{p}). By using conformal prediction, we can obtain a safer and more trustworthy predictive set for MSSA 3 since the top-2 predictions Group C Strep. and Group A Strep. require different antibiotics (Penicillian) than MSSA 3 (Vancomycin)\footnotemark}
   \label{fig:conformal_prediction_result}
\end{figure*}

To demonstrate the benefit of using conformal prediction, we examine the relation between emperical coverage, theoretical coverage, and average prediction set size in Fig.~\ref{fig:conformal_prediction_result}. The average set size is the average of the length of the predictive sets over all the test spectra. A smaller average set size conveys more detailed information and may be more useful in practice since it will take less time to examine each of the predictions from the set. 

We observe that as the theoretical coverage increases, the empirical coverage also increases with a moderate increase of the average set size. Compared to using a fixed-size top-$M$ predicting strategy, we can achieve a much higher matching accuracy by increasing the set size using conformal prediction. For example, with an average set size of 2 we increase our accuracy from 83\% to 90\% and with an average set size of 7, we can correctly identify 99\% of the bacteria. 

\footnotetext{The information regarding the treatment for the bacteria is provided by~\citet{Ho2019RapidIO}}

\section{Conclusion}
In this paper, we proposed to use a Siamese network for Raman spectrum matching. Given pairs of spectra as input, the model learns a feature representation that is optimized for matching spectra belonging to the same class. We use an Xception-based architecture in our network that allows the extracted feature to contain the fingerprint characteristics such as locations, shapes, and the number of peaks as much as possible while keeping the model complexity low. We experimentally demonstrate that we match or outperform existing approaches over three publicly available datasets with a similar or much smaller number of parameters. Our method does not require delicate preprocessing steps such as baseline correction, smoothing, or normalization: In fact, it gives the best results on raw data. The model also reliably estimates the uncertainty of the predictions which is useful to signal when the model is likely to make a wrong prediction. We further use a conformal prediction technique to compute prediction sets that cover the true class with a user-defined coverage level (i.e., 99\%). This technique is especially critical for sensitive tasks (i.e., the detection of toxic or explosive materials using Raman spectroscopy).

In practical application outside the laboratory, the measured Raman spectra rarly only include a single component. Therefore, our future work will investigate the spectrum matching for mixture components.

\begin{acknowledgement}
The authors thank for financial support from the European Union’s Horizon 2020 research and innovation programme under grant agreement no. 883390 (H2020-SU-SECU-2019 SERSing Project). The authors thank the NVIDIA Corporation with the donation of GPUs used for this research. The authors thank Sebastian Farås, Serstech AB, Sweden, for the fruitful discussion concerning spectral matching.

\end{acknowledgement}

\newpage
\begin{suppinfo}

\renewcommand{\thesection}{S}
\begin{itemize}
    \item S.1: Distribution of the spectra per class in the Mineral dataset
    \item S.2: Influence of different augmentation methods
    \item S.3: Influence of the positive-negative ratio 
    \item S.4: Influence of similarity matching criteria
    \item S.5: Error analysis on S.lugdunensis in the Bacteria dataset
\end{itemize}


\end{suppinfo}

\bibliography{ref.bib}

\providecommand{\latin}[1]{#1}
\makeatletter
\providecommand{\doi}
  {\begingroup\let\do\@makeother\dospecials
  \catcode`\{=1 \catcode`\}=2 \doi@aux}
\providecommand{\doi@aux}[1]{\endgroup\texttt{#1}}
\makeatother
\providecommand*\mcitethebibliography{\thebibliography}
\csname @ifundefined\endcsname{endmcitethebibliography}
  {\let\endmcitethebibliography\endthebibliography}{}
\begin{mcitethebibliography}{36}
\providecommand*\natexlab[1]{#1}
\providecommand*\mciteSetBstSublistMode[1]{}
\providecommand*\mciteSetBstMaxWidthForm[2]{}
\providecommand*\mciteBstWouldAddEndPuncttrue
  {\def\EndOfBibitem{\unskip.}}
\providecommand*\mciteBstWouldAddEndPunctfalse
  {\let\EndOfBibitem\relax}
\providecommand*\mciteSetBstMidEndSepPunct[3]{}
\providecommand*\mciteSetBstSublistLabelBeginEnd[3]{}
\providecommand*\EndOfBibitem{}
\mciteSetBstSublistMode{f}
\mciteSetBstMaxWidthForm{subitem}{(\alph{mcitesubitemcount})}
\mciteSetBstSublistLabelBeginEnd
  {\mcitemaxwidthsubitemform\space}
  {\relax}
  {\relax}

\bibitem[Amjad \latin{et~al.}(2018)Amjad, Ullah, Khan, Bilal, and
  Khan]{AMJAD2018124}
Amjad,~A.; Ullah,~R.; Khan,~S.; Bilal,~M.; Khan,~A. Raman spectroscopy based
  analysis of milk using random forest classification. \emph{Vibrational
  Spectroscopy} \textbf{2018}, \emph{99}, 124--129\relax
\mciteBstWouldAddEndPuncttrue
\mciteSetBstMidEndSepPunct{\mcitedefaultmidpunct}
{\mcitedefaultendpunct}{\mcitedefaultseppunct}\relax
\EndOfBibitem
\bibitem[Lussier \latin{et~al.}(2020)Lussier, Thibault, Charron, Wallace, and
  Masson]{lussier2020a}
Lussier,~F.; Thibault,~V.; Charron,~B.; Wallace,~G.~Q.; Masson,~J.~F. Deep
  learning and artificial intelligence methods for Raman and surface-enhanced
  Raman scattering. \emph{Trac - Trends in Analytical Chemistry} \textbf{2020},
  \emph{124}, 115796\relax
\mciteBstWouldAddEndPuncttrue
\mciteSetBstMidEndSepPunct{\mcitedefaultmidpunct}
{\mcitedefaultendpunct}{\mcitedefaultseppunct}\relax
\EndOfBibitem
\bibitem[Huang and Yu(2021)Huang, and Yu]{huang2021a}
Huang,~T.~Y.; Yu,~J. C.~C. Development of crime scene intelligence using a
  hand-held raman spectrometer and transfer learning. \emph{Analytical
  Chemistry} \textbf{2021}, \emph{93}, 8889--8896\relax
\mciteBstWouldAddEndPuncttrue
\mciteSetBstMidEndSepPunct{\mcitedefaultmidpunct}
{\mcitedefaultendpunct}{\mcitedefaultseppunct}\relax
\EndOfBibitem
\bibitem[Ho \latin{et~al.}(2019)Ho, Jean, Hogan, Blackmon, Jeffrey, Holodniy,
  Banaei, Saleh, Ermon, and Dionne]{Ho2019RapidIO}
Ho,~C.-S.; Jean,~N.; Hogan,~C.; Blackmon,~L.; Jeffrey,~S.; Holodniy,~M.;
  Banaei,~N.; Saleh,~A. A.~E.; Ermon,~S.; Dionne,~J. Rapid identification of
  pathogenic bacteria using Raman spectroscopy and deep learning. \emph{Nature
  Communications} \textbf{2019}, \emph{10}\relax
\mciteBstWouldAddEndPuncttrue
\mciteSetBstMidEndSepPunct{\mcitedefaultmidpunct}
{\mcitedefaultendpunct}{\mcitedefaultseppunct}\relax
\EndOfBibitem
\bibitem[Liu \latin{et~al.}(2019)Liu, Gibson, Mills, and Osadchy]{LIU2019175}
Liu,~J.; Gibson,~S.~J.; Mills,~J.; Osadchy,~M. Dynamic spectrum matching with
  one-shot learning. \emph{Chemometrics and Intelligent Laboratory Systems}
  \textbf{2019}, \emph{184}, 175--181\relax
\mciteBstWouldAddEndPuncttrue
\mciteSetBstMidEndSepPunct{\mcitedefaultmidpunct}
{\mcitedefaultendpunct}{\mcitedefaultseppunct}\relax
\EndOfBibitem
\bibitem[Lafuente \latin{et~al.}(2016)Lafuente, Downs, Yang, and
  Stone]{Lafuente2016ThePO}
Lafuente,~B.; Downs,~R.; Yang,~H.; Stone,~N. The power of databases: The RRUFF
  project. 2016\relax
\mciteBstWouldAddEndPuncttrue
\mciteSetBstMidEndSepPunct{\mcitedefaultmidpunct}
{\mcitedefaultendpunct}{\mcitedefaultseppunct}\relax
\EndOfBibitem
\bibitem[Liu \latin{et~al.}(2017)Liu, Osadchy, Ashton, Foster, Solomon, and
  Gibson]{C7AN01371J}
Liu,~J.; Osadchy,~M.; Ashton,~L.; Foster,~M.; Solomon,~C.~J.; Gibson,~S.~J.
  Deep convolutional neural networks for Raman spectrum recognition: a unified
  solution. \emph{Analyst} \textbf{2017}, \emph{142}, 4067--4074\relax
\mciteBstWouldAddEndPuncttrue
\mciteSetBstMidEndSepPunct{\mcitedefaultmidpunct}
{\mcitedefaultendpunct}{\mcitedefaultseppunct}\relax
\EndOfBibitem
\bibitem[McLaughlin \latin{et~al.}(2014)McLaughlin, Doty, and
  Lednev]{doi:10.1021/ac5026368}
McLaughlin,~G.; Doty,~K.~C.; Lednev,~I.~K. Raman Spectroscopy of Blood for
  Species Identification. \emph{Analytical Chemistry} \textbf{2014}, \emph{86},
  11628--11633\relax
\mciteBstWouldAddEndPuncttrue
\mciteSetBstMidEndSepPunct{\mcitedefaultmidpunct}
{\mcitedefaultendpunct}{\mcitedefaultseppunct}\relax
\EndOfBibitem
\bibitem[de~Juan \latin{et~al.}(2014)de~Juan, Jaumot, and Tauler]{C4AY00571F}
de~Juan,~A.; Jaumot,~J.; Tauler,~R. Multivariate Curve Resolution (MCR).
  Solving the mixture analysis problem. \emph{Anal. Methods} \textbf{2014},
  \emph{6}, 4964--4976\relax
\mciteBstWouldAddEndPuncttrue
\mciteSetBstMidEndSepPunct{\mcitedefaultmidpunct}
{\mcitedefaultendpunct}{\mcitedefaultseppunct}\relax
\EndOfBibitem
\bibitem[Alstrøm \latin{et~al.}(2014)Alstrøm, Frøhling, Larsen, Schmidt,
  Bache, Schmidt, Jakobsen, and Boisen]{6958925}
Alstrøm,~T.~S.; Frøhling,~K.~B.; Larsen,~J.; Schmidt,~M.~N.; Bache,~M.;
  Schmidt,~M.~S.; Jakobsen,~M.~H.; Boisen,~A. Improving the robustness of
  Surface Enhanced Raman Spectroscopy based sensors by Bayesian Non-negative
  Matrix Factorization. 2014 IEEE International Workshop on Machine Learning
  for Signal Processing (MLSP). 2014; pp 1--6\relax
\mciteBstWouldAddEndPuncttrue
\mciteSetBstMidEndSepPunct{\mcitedefaultmidpunct}
{\mcitedefaultendpunct}{\mcitedefaultseppunct}\relax
\EndOfBibitem
\bibitem[Virkler and Lednev(2009)Virkler, and Lednev]{doi:10.1021/ac901350a}
Virkler,~K.; Lednev,~I.~K. Blood Species Identification for Forensic Purposes
  Using Raman Spectroscopy Combined with Advanced Statistical Analysis.
  \emph{Analytical Chemistry} \textbf{2009}, \emph{81}, 7773--7777\relax
\mciteBstWouldAddEndPuncttrue
\mciteSetBstMidEndSepPunct{\mcitedefaultmidpunct}
{\mcitedefaultendpunct}{\mcitedefaultseppunct}\relax
\EndOfBibitem
\bibitem[Hu \latin{et~al.}(2019)Hu, Ye, Zhang, Li, Zhang, Luo, Mukamel, and
  Jiang]{doi:10.1021/acs.jpclett.9b02517}
Hu,~W.; Ye,~S.; Zhang,~Y.; Li,~T.; Zhang,~G.; Luo,~Y.; Mukamel,~S.; Jiang,~J.
  Machine Learning Protocol for Surface-Enhanced Raman Spectroscopy. \emph{The
  Journal of Physical Chemistry Letters} \textbf{2019}, \emph{10},
  6026--6031\relax
\mciteBstWouldAddEndPuncttrue
\mciteSetBstMidEndSepPunct{\mcitedefaultmidpunct}
{\mcitedefaultendpunct}{\mcitedefaultseppunct}\relax
\EndOfBibitem
\bibitem[Sevetlidis and Pavlidis(2019)Sevetlidis, and
  Pavlidis]{SEVETLIDIS2019121}
Sevetlidis,~V.; Pavlidis,~G. Effective Raman spectra identification with
  tree-based methods. \emph{Journal of Cultural Heritage} \textbf{2019},
  \emph{37}, 121--128\relax
\mciteBstWouldAddEndPuncttrue
\mciteSetBstMidEndSepPunct{\mcitedefaultmidpunct}
{\mcitedefaultendpunct}{\mcitedefaultseppunct}\relax
\EndOfBibitem
\bibitem[Kang \latin{et~al.}(2021)Kang, Kim, and
  Vikesland]{doi:10.1021/acs.analchem.0c04576}
Kang,~S.; Kim,~I.; Vikesland,~P.~J. Discriminatory Detection of ssDNA by
  Surface-Enhanced Raman Spectroscopy (SERS) and Tree-Based Support Vector
  Machine (Tr-SVM). \emph{Analytical Chemistry} \textbf{2021}, \emph{93},
  9319--9328\relax
\mciteBstWouldAddEndPuncttrue
\mciteSetBstMidEndSepPunct{\mcitedefaultmidpunct}
{\mcitedefaultendpunct}{\mcitedefaultseppunct}\relax
\EndOfBibitem
\bibitem[Fukuhara \latin{et~al.}(2019)Fukuhara, Fujiwara, Maruyama, and
  Itoh]{FUKUHARA201911}
Fukuhara,~M.; Fujiwara,~K.; Maruyama,~Y.; Itoh,~H. Feature visualization of
  Raman spectrum analysis with deep convolutional neural network.
  \emph{Analytica Chimica Acta} \textbf{2019}, \emph{1087}, 11--19\relax
\mciteBstWouldAddEndPuncttrue
\mciteSetBstMidEndSepPunct{\mcitedefaultmidpunct}
{\mcitedefaultendpunct}{\mcitedefaultseppunct}\relax
\EndOfBibitem
\bibitem[Chollet(2017)]{DBLP:conf/cvpr/Chollet17}
Chollet,~F. Xception: Deep Learning with Depthwise Separable Convolutions. 2017
  {IEEE} Conference on Computer Vision and Pattern Recognition, {CVPR} 2017,
  Honolulu, HI, USA, July 21-26, 2017. 2017; pp 1800--1807\relax
\mciteBstWouldAddEndPuncttrue
\mciteSetBstMidEndSepPunct{\mcitedefaultmidpunct}
{\mcitedefaultendpunct}{\mcitedefaultseppunct}\relax
\EndOfBibitem
\bibitem[He \latin{et~al.}(2016)He, Zhang, Ren, and
  Sun]{DBLP:conf/cvpr/HeZRS16}
He,~K.; Zhang,~X.; Ren,~S.; Sun,~J. Deep Residual Learning for Image
  Recognition. 2016 {IEEE} Conference on Computer Vision and Pattern
  Recognition, {CVPR} 2016, Las Vegas, NV, USA, June 27-30, 2016. 2016; pp
  770--778\relax
\mciteBstWouldAddEndPuncttrue
\mciteSetBstMidEndSepPunct{\mcitedefaultmidpunct}
{\mcitedefaultendpunct}{\mcitedefaultseppunct}\relax
\EndOfBibitem
\bibitem[Lakshminarayanan \latin{et~al.}(2017)Lakshminarayanan, Pritzel, and
  Blundell]{DBLP:conf/nips/Lakshminarayanan17}
Lakshminarayanan,~B.; Pritzel,~A.; Blundell,~C. Simple and Scalable Predictive
  Uncertainty Estimation using Deep Ensembles. Advances in Neural Information
  Processing Systems 30: Annual Conference on Neural Information Processing
  Systems 2017, December 4-9, 2017, Long Beach, CA, {USA}. 2017; pp
  6402--6413\relax
\mciteBstWouldAddEndPuncttrue
\mciteSetBstMidEndSepPunct{\mcitedefaultmidpunct}
{\mcitedefaultendpunct}{\mcitedefaultseppunct}\relax
\EndOfBibitem
\bibitem[Yu \latin{et~al.}(2021)Yu, Li, Lu, Li, Fu, and
  Liu]{doi:10.1021/acs.analchem.1c00431}
Yu,~S.; Li,~X.; Lu,~W.; Li,~H.; Fu,~Y.~V.; Liu,~F. Analysis of Raman Spectra by
  Using Deep Learning Methods in the Identification of Marine Pathogens.
  \emph{Analytical Chemistry} \textbf{2021}, \emph{93}, 11089--11098\relax
\mciteBstWouldAddEndPuncttrue
\mciteSetBstMidEndSepPunct{\mcitedefaultmidpunct}
{\mcitedefaultendpunct}{\mcitedefaultseppunct}\relax
\EndOfBibitem
\bibitem[Schmidt \latin{et~al.}(2019)Schmidt, Alstr{\o}m, Svendstorp, and
  Larsen]{schmidt2019icassp}
Schmidt,~M.~N.; Alstr{\o}m,~T.~S.; Svendstorp,~M.; Larsen,~J. Peak detection
  and baseline correction using a convolutional neural network. Acoustics,
  speech and signal processing, IEEE international conference on (ICASSP).
  2019\relax
\mciteBstWouldAddEndPuncttrue
\mciteSetBstMidEndSepPunct{\mcitedefaultmidpunct}
{\mcitedefaultendpunct}{\mcitedefaultseppunct}\relax
\EndOfBibitem
\bibitem[Guo \latin{et~al.}(2017)Guo, Pleiss, Sun, and Weinberger]{Guo2017}
Guo,~C.; Pleiss,~G.; Sun,~Y.; Weinberger,~K.~Q. On calibration of modern neural
  networks. 2017\relax
\mciteBstWouldAddEndPuncttrue
\mciteSetBstMidEndSepPunct{\mcitedefaultmidpunct}
{\mcitedefaultendpunct}{\mcitedefaultseppunct}\relax
\EndOfBibitem
\bibitem[Angelopoulos \latin{et~al.}(2021)Angelopoulos, Bates, Jordan, and
  Malik]{angelopoulos2021uncertainty}
Angelopoulos,~A.~N.; Bates,~S.; Jordan,~M.; Malik,~J. Uncertainty Sets for
  Image Classifiers using Conformal Prediction. International Conference on
  Learning Representations. 2021\relax
\mciteBstWouldAddEndPuncttrue
\mciteSetBstMidEndSepPunct{\mcitedefaultmidpunct}
{\mcitedefaultendpunct}{\mcitedefaultseppunct}\relax
\EndOfBibitem
\bibitem[Romano \latin{et~al.}(2020)Romano, Sesia, and
  Cand{\`{e}}s]{DBLP:conf/nips/RomanoSC20}
Romano,~Y.; Sesia,~M.; Cand{\`{e}}s,~E.~J. Classification with Valid and
  Adaptive Coverage. Advances in Neural Information Processing Systems 33:
  Annual Conference on Neural Information Processing Systems 2020, NeurIPS
  2020, December 6-12, 2020, virtual. 2020\relax
\mciteBstWouldAddEndPuncttrue
\mciteSetBstMidEndSepPunct{\mcitedefaultmidpunct}
{\mcitedefaultendpunct}{\mcitedefaultseppunct}\relax
\EndOfBibitem
\bibitem[Zhang \latin{et~al.}(2020)Zhang, Xie, Cai, Hu, Liu, Hong, and
  Tian]{organic_dataset}
Zhang,~R.; Xie,~H.; Cai,~S.; Hu,~Y.; Liu,~G.-k.; Hong,~W.; Tian,~Z.-q.
  Transfer-learning-based Raman spectra identification. \emph{Journal of Raman
  Spectroscopy} \textbf{2020}, \emph{51}, 176--186\relax
\mciteBstWouldAddEndPuncttrue
\mciteSetBstMidEndSepPunct{\mcitedefaultmidpunct}
{\mcitedefaultendpunct}{\mcitedefaultseppunct}\relax
\EndOfBibitem
\bibitem[Koch(2015)]{Koch2015SiameseNN}
Koch,~G.~R. Siamese Neural Networks for One-Shot Image Recognition. 2015\relax
\mciteBstWouldAddEndPuncttrue
\mciteSetBstMidEndSepPunct{\mcitedefaultmidpunct}
{\mcitedefaultendpunct}{\mcitedefaultseppunct}\relax
\EndOfBibitem
\bibitem[Szegedy \latin{et~al.}(2017)Szegedy, Ioffe, Vanhoucke, and
  Alemi]{DBLP:conf/aaai/SzegedyIVA17}
Szegedy,~C.; Ioffe,~S.; Vanhoucke,~V.; Alemi,~A.~A. Inception-v4,
  Inception-ResNet and the Impact of Residual Connections on Learning.
  Proceedings of the Thirty-First {AAAI} Conference on Artificial Intelligence,
  February 4-9, 2017, San Francisco, California, {USA}. 2017; pp
  4278--4284\relax
\mciteBstWouldAddEndPuncttrue
\mciteSetBstMidEndSepPunct{\mcitedefaultmidpunct}
{\mcitedefaultendpunct}{\mcitedefaultseppunct}\relax
\EndOfBibitem
\bibitem[Szegedy \latin{et~al.}(2015)Szegedy, Liu, Jia, Sermanet, Reed,
  Anguelov, Erhan, Vanhoucke, and Rabinovich]{DBLP:conf/cvpr/SzegedyLJSRAEVR15}
Szegedy,~C.; Liu,~W.; Jia,~Y.; Sermanet,~P.; Reed,~S.~E.; Anguelov,~D.;
  Erhan,~D.; Vanhoucke,~V.; Rabinovich,~A. Going deeper with convolutions.
  {IEEE} Conference on Computer Vision and Pattern Recognition, {CVPR} 2015,
  Boston, MA, USA, June 7-12, 2015. 2015; pp 1--9\relax
\mciteBstWouldAddEndPuncttrue
\mciteSetBstMidEndSepPunct{\mcitedefaultmidpunct}
{\mcitedefaultendpunct}{\mcitedefaultseppunct}\relax
\EndOfBibitem
\bibitem[Gotmare \latin{et~al.}(2019)Gotmare, Keskar, Xiong, and
  Socher]{DBLP:conf/iclr/GotmareKXS19}
Gotmare,~A.; Keskar,~N.~S.; Xiong,~C.; Socher,~R. A Closer Look at Deep
  Learning Heuristics: Learning rate restarts, Warmup and Distillation. 7th
  International Conference on Learning Representations, {ICLR} 2019, New
  Orleans, LA, USA, May 6-9, 2019. 2019\relax
\mciteBstWouldAddEndPuncttrue
\mciteSetBstMidEndSepPunct{\mcitedefaultmidpunct}
{\mcitedefaultendpunct}{\mcitedefaultseppunct}\relax
\EndOfBibitem
\bibitem[Dietterich(2000)]{DBLP:conf/mcs/Dietterich00}
Dietterich,~T.~G. Ensemble Methods in Machine Learning. Multiple Classifier
  Systems, First International Workshop, {MCS} 2000, Cagliari, Italy, June
  21-23, 2000, Proceedings. 2000; pp 1--15\relax
\mciteBstWouldAddEndPuncttrue
\mciteSetBstMidEndSepPunct{\mcitedefaultmidpunct}
{\mcitedefaultendpunct}{\mcitedefaultseppunct}\relax
\EndOfBibitem
\bibitem[Candela \latin{et~al.}(2005)Candela, Rasmussen, Sinz, Bousquet, and
  Sch{\"{o}}lkopf]{DBLP:conf/mlcw/CandelaRSBS05}
Candela,~J.~Q.; Rasmussen,~C.~E.; Sinz,~F.~H.; Bousquet,~O.;
  Sch{\"{o}}lkopf,~B. Evaluating Predictive Uncertainty Challenge. Machine
  Learning Challenges, Evaluating Predictive Uncertainty, Visual Object
  Classification and Recognizing Textual Entailment, First {PASCAL} Machine
  Learning Challenges Workshop, {MLCW} 2005, Southampton, UK, April 11-13,
  2005, Revised Selected Papers. 2005; pp 1--27\relax
\mciteBstWouldAddEndPuncttrue
\mciteSetBstMidEndSepPunct{\mcitedefaultmidpunct}
{\mcitedefaultendpunct}{\mcitedefaultseppunct}\relax
\EndOfBibitem
\bibitem[Lee \latin{et~al.}(2015)Lee, Purushwalkam, Cogswell, Crandall, and
  Batra]{DBLP:journals/corr/LeePCCB15}
Lee,~S.; Purushwalkam,~S.; Cogswell,~M.; Crandall,~D.~J.; Batra,~D. Why M Heads
  are Better than One: Training a Diverse Ensemble of Deep Networks.
  \emph{CoRR} \textbf{2015}, \emph{abs/1511.06314}\relax
\mciteBstWouldAddEndPuncttrue
\mciteSetBstMidEndSepPunct{\mcitedefaultmidpunct}
{\mcitedefaultendpunct}{\mcitedefaultseppunct}\relax
\EndOfBibitem
\bibitem[Vovk \latin{et~al.}(2005)Vovk, Gammerman, and Shafer]{Vovk20051}
Vovk,~V.; Gammerman,~A.; Shafer,~G. \emph{Algorithmic learning in a random
  world}; 2005; pp 1--324, cited By 521\relax
\mciteBstWouldAddEndPuncttrue
\mciteSetBstMidEndSepPunct{\mcitedefaultmidpunct}
{\mcitedefaultendpunct}{\mcitedefaultseppunct}\relax
\EndOfBibitem
\bibitem[Shafer and Vovk(2008)Shafer, and Vovk]{DBLP:journals/jmlr/ShaferV08}
Shafer,~G.; Vovk,~V. A Tutorial on Conformal Prediction. \emph{J. Mach. Learn.
  Res.} \textbf{2008}, \emph{9}, 371--421\relax
\mciteBstWouldAddEndPuncttrue
\mciteSetBstMidEndSepPunct{\mcitedefaultmidpunct}
{\mcitedefaultendpunct}{\mcitedefaultseppunct}\relax
\EndOfBibitem
\bibitem[Papadopoulos \latin{et~al.}(2002)Papadopoulos, Proedrou, Vovk, and
  Gammerman]{DBLP:conf/ecml/PapadopoulosPVG02}
Papadopoulos,~H.; Proedrou,~K.; Vovk,~V.; Gammerman,~A. Inductive Confidence
  Machines for Regression. Machine Learning: {ECML} 2002, 13th European
  Conference on Machine Learning, Helsinki, Finland, August 19-23, 2002,
  Proceedings. 2002; pp 345--356\relax
\mciteBstWouldAddEndPuncttrue
\mciteSetBstMidEndSepPunct{\mcitedefaultmidpunct}
{\mcitedefaultendpunct}{\mcitedefaultseppunct}\relax
\EndOfBibitem
\bibitem[Papadopoulos \latin{et~al.}(2007)Papadopoulos, Vovk, and
  Gammerman]{DBLP:conf/ictai/PapadopoulosVG07}
Papadopoulos,~H.; Vovk,~V.; Gammerman,~A. Conformal Prediction with Neural
  Networks. 19th {IEEE} International Conference on Tools with Artificial
  Intelligence {(ICTAI} 2007), October 29-31, 2007, Patras, Greece, Volume 2.
  2007; pp 388--395\relax
\mciteBstWouldAddEndPuncttrue
\mciteSetBstMidEndSepPunct{\mcitedefaultmidpunct}
{\mcitedefaultendpunct}{\mcitedefaultseppunct}\relax
\EndOfBibitem
\end{mcitethebibliography}

\end{document}



\subsection{Dataset statistics}
\label{sec:sup_data}
Fig.~\ref{fig:unbalanced_class_distribution_on_RRUFF} shows the distribution of spectra per class in the Mineral dataset. The majority of the minerals (50\%) have less than five spectra.
\begin{figure}[ht!]
    \centering
    \importpgf{PGF_figures/class_distribution_on_RRUFF.pgf}
    \caption[]{The distribution of spectra per class in the Mineral dataset.}
    \label{fig:unbalanced_class_distribution_on_RRUFF}
\end{figure}

\subsection{Influence of the augmentation methods}
\label{sec:sup_augmentation}
We also evaluate the matching accuracy on the Mineral dataset using different augmentation methods. Since it is common to generate the synthetic ones by taking the interpolation with randomly simulated coefficients if the number of spectra is larger than one~\cite{C7AN01371J, LIU2019175, organic_dataset}, we here only experiment with applying different augmentation methods on the classes that only have one spectrum in the training data. We choose to use different augmentation methods as described below:
\begin{itemize}
    \item None: no augmentation is applied
    \item Repeat: we duplicate the spectrum for each class to a certain number
    \item G-noise: we apply randomly generated Gaussian noise 
    \item D-noise: we simulate noise based on the variation of the first derivative of the intensities as described in Section~2 and add those to the original spectrum
\end{itemize}
We experiment with different augmentation methods using the same data splits and we repeat this process five times. The results are shown in Table~\ref{tab:augmentatation_results_on_RRUFF}. The model is relatively more stable when we apply augmentations as described in section~2. Besides, we also achieved a slightly higher accuracy. 

\begin{table}[ht!]
\caption{Matching accuracy with 95\% confidence interval for the Mineral dataset using different augmentations. Generating synthetic spectra with \textit{D-noise} leads to a more stable and slightly higher spectrum matching accuracy.}
\label{tab:augmentatation_results_on_RRUFF}
\begin{tabular}{lcccc}
\hline
 & None & Repeat & G-noise & D-noise \\ \hline \noalign{\smallskip}
Mineral (r) & 93.58$\pm$0.68 & 93.94$\pm$0.54 & 93.64$\pm$0.32 & \textbf{94.05$\pm$0.25} \\ \noalign{\smallskip}
Mineral (p) & 91.50$\pm$0.25 & 91.58$\pm$0.38 & 91.42$\pm$0.55 & \textbf{92.34$\pm$0.26}\\ \bottomrule
\end{tabular}
\end{table}

\subsection{Influence of the positive-negative ratio}
\label{sec:sup_ratio}
One of the problems in training a Siamese network is how to balance the number of positive pairs and negative pairs per batch $\alpha$ such that the model is able to retrieve the similar ones and also discriminate the dissimilar ones. Therefore, we here examine the influence of the ratio $\alpha$ on the model performance for all the dataset on a fixed data splits. The results are shown in Table.~\ref{tab:influece_of_pos_neg_ratio_allinone}. 

\begin{table}[ht!]
\caption{Influence of the ratio between the number of positive pairs and negative pairs per batch $\alpha$ over five different data splits ($\pm$95\% confidence interval). The choice of the $\alpha$ has more visible influence on the Bacteria dataset compared to other datasets}
\smallskip
\label{tab:influece_of_pos_neg_ratio_allinone}
\centering
\begin{tabular}{@{}lccccc@{}}
\toprule
 & 0.05 & 0.1 & 0.5 & 1 & 2 \\ \midrule
Mineral (r) & 93.84±0.49 & 93.74±0.64 & 94.20±0.67 & \textbf{94.66±0.55} & 94.05±0.47 \\
Mineral (p) & 91.73±0.16 & 92.09±0.28 & 92.22±0.37 & \textbf{92.24±0.32} & 92.19±0.27 \\
Organic (r) & 94.44±0.76 & 96.11±1.19 & \textbf{97.50±0.91} & 97.22±0.77 & 96.66±0.97 \\
Organic (p) & \textbf{97.22±0.77} & \textbf{97.22±0.77} & \textbf{97.22±0.77} & 97.22±1.33 & 96.66±0.97 \\
Bacteria & 82.44±0.39 & 82.28±0.45 & \textbf{83.06±0.19} & 82.14±0.48 & 82.12±0.40 \\ \bottomrule
\end{tabular}
\end{table}


\subsection{Influence of the distance calculation} 
\label{sec:sup_distance}
To demonstrate the need of using both the element-wise product and the absolute difference between the feature maps as input to the neural network that computes the spectral similarity, we conduct a study where we drop one of the distance metrics. The results over multiple data splits are shown in the Table~\ref{tab:distance_results_on_all_datasets}. Absolute difference alone performs better than the element-wise product on the Mineral (r) and Organic datasets, but worse than the element-wise product on the Bacteria dataset. Using both metrics together gives better and more stable performance on all the datasets. 

\begin{table}[ht!]
\caption{Matching accuracy with 95\% confidence interval over five data splits on all the datasets.}
\label{tab:distance_results_on_all_datasets}
\begin{tabular}{@{}lccc@{}}
\toprule
 & $D_\text{prod.}$ & $D_\text{diff.}$ & $D_\text{prod.}$ and $D_\text{diff.}$\\ \midrule
Mineral (r) & 94.20$\pm$0.39 & 94.25$\pm$0.81 & \textbf{94.51$\pm$0.15} \\
Mineral (p) & \textbf{92.25$\pm$0.49} & 90.23$\pm$0.36 & 91.85$\pm$0.34 \\
Organic (r) & 92.78$\pm$3.20 & 96.94$\pm$2.35 & \textbf{96.39$\pm$0.59} \\
Organic (p) & 94.21$\pm$1.18 & 96.52$\pm$1.06 & \textbf{96.76$\pm$1.38} \\
Bacteria & 82.22$\pm$0.72 & 81.13$\pm$ 0.34 & \textbf{82.63$\pm$0.32}  \\ \bottomrule
\end{tabular}
\end{table}

\subsection{Error analysis on S. lugdunensis}
\label{sec:sup_slugdunensis}
S. Lugdunensis is an example of a class in the Bacteria dataset that is difficult to classify correctly using spectral matching, because the spectrum in the test dataset is different than the ones in the reference and reference-finetune dataset (see Fig.~\ref{fig:bacteria_trouble_maker}.)
\begin{figure}[ht!]
    \centering
    \importpgf{PGF_figures/bacteria_troublemaker.pgf}
    \caption{Averaged spectrum for Bacteria \textit{S. lugdunensis} and \textit{Group C Strep.} in the reference, reference-finetune, and the test dataset. }
    \label{fig:bacteria_trouble_maker}
\end{figure}